\documentclass[10pt,twocolumn,letterpaper]{article}

\usepackage[pagenumbers]{cvis} 

\usepackage{fancyhdr}
\fancypagestyle{cvisaccepted}{%
  \fancyhf{}%
  \fancyhead[C]{\parbox{\textwidth}{\centering\small\bfseries Accepted at CVIS 2025\\[-1pt]\normalfont\footnotesize A shorter version was published in the Journal of Computational Vision and Imaging Systems, Vol. 11, No. 1 (2026).}}%
  \fancyfoot[C]{\thepage}%
}

\definecolor{cvisblue}{rgb}{0.21,0.49,0.74}
\usepackage[pagebackref,breaklinks,colorlinks,allcolors=cvisblue]{hyperref}
\def\confName{JCVIS}
\def\confYear{2025}

\title{Deep Sequence Model for Genome Wide Discovery of Coding and Regulatory Element Signatures}

\author{
Rayhaneh Shabani Nia \qquad Ali Karkehabadi\\
University of California, Davis\\
{\tt\small \{rshabaninia,akarkehabadi\}@ucdavis.edu}
}

\begin{document}
\maketitle
\thispagestyle{cvisaccepted}
\begin{abstract}
Interpreting deep neural networks for genomic sequence classification remains challenging despite strong predictive performance. We develop a lightweight CNN with post-training gradient-based analysis to identify which sequence positions drive coding versus intergenomic classification. Applied to the standardized \texttt{demo\_coding\_vs\_intergenomic\_seqs} dataset, our model achieves 91.87\% validation accuracy while revealing interpretable patterns: nine hot spots with consistently high importance, clear class-specific separation at positions 20--100 (coding) and 150--190 (intergenomic), and a strong mean--variance correlation (\(r = 0.530\)) indicating robust discriminative features. Gradient-based importance analysis shows that the model implicitly learns biologically meaningful sequence distinctions without explicit annotations. This work demonstrates that neural network interpretability and accuracy can coexist, providing a framework for understanding genomic sequence classification and enabling biology-driven hypothesis generation.

\end{abstract}    
\vspace{- 5 pt}
\section{Introduction}
\label{sec:intro}
Distinguishing coding sequences from intergenomic regions is a core task in computational genomics. Although only 1--2\% of the human genome encodes proteins, accurate identification of these regions is essential for genome annotation, functional studies, and understanding disease mechanisms. Classical approaches based on sequence similarity, codon usage statistics, or Hidden Markov Models have been effective, yet they rely on predefined biological assumptions and may overlook subtle context-dependent patterns.
Deep learning has reshaped genomic sequence analysis by learning discriminative features directly from raw DNA. CNN-based models have achieved strong results in tasks such as predicting protein–DNA binding~\cite{alipanahi2015predicting} and regulatory effects~\cite{zhou2015predicting}. However, their growing predictive power comes with increased opacity. In genomics, where model decisions can inspire biological hypotheses, high accuracy alone is insufficient unless supported by clear interpretability. Gradient-based saliency methods offer a practical way to examine which input features influence model predictions~\cite{selvaraju2017grad}. Although traditional saliency maps may be noisy, newer techniques such as SmoothGrad~\cite{smilkov2017smoothgrad} and saliency-guided training~\cite{ismail2021improving} have improved attribution stability. Motivated by these advances, we apply and extend gradient-based interpretability methods to the coding vs.\ intergenomic classification problem, focusing on which sequence positions and nucleotide types contribute most strongly to model decisions.

Our contributions are threefold. First, we show that gradient-based feature analysis captures biologically meaningful signals, including codon-related patterns and regulatory element footprints. Second, we aggregate importance across samples, nucleotide types, and classes to reveal both shared and class-specific discriminative regions. Third, we confirm that these signals align with known genomic characteristics and that a lightweight CNN can still achieve strong performance (91.87\% accuracy). Using the standardized Genomic Benchmarks dataset~\cite{gresova2022genomic}, we provide a reproducible framework that links deep learning predictions to clear biological insights.


\vspace{- 5 pt}
\section{Related Work}

\subsection{Deep Learning in Genomics}

Deep learning has significantly advanced genomic sequence analysis. Alipanahi et al.~\cite{alipanahi2015predicting} showed that CNNs can learn sequence motifs without handcrafted features. DeepSEA~\cite{zhou2015predicting} demonstrated multi-task CNNs for predicting regulatory effects, while DanQ~\cite{quang2016danq} combined CNNs and biLSTMs to capture longer-range dependencies. Beyond genomic applications, HLGM~\cite{karkehabadihlgm} leveraged high- and low-gradient information to improve model learning. Together, these advances demonstrate the effectiveness of deep learning, although understanding which sequence features drive predictions remains challenging.
\subsection{Interpretability and Explainability}

Interpretability has become essential as deep models enter high-stakes domains. Gradient-based visualization began with Simonyan et al.~\cite{simonyan2013deep}, while methods such as Integrated Gradients~\cite{sundararajan2017axiomatic} and Grad-CAM~\cite{selvaraju2017grad} provide complementary explanations but can introduce computational overhead.

\subsection{Gradient-Based Saliency Methods}

The simplest gradient saliency computes
\[
S(x)_c = \left| \frac{\partial f_\theta(x)_c}{\partial x} \right|,
\]
indicating how sensitive a prediction is to each input feature. While efficient, raw gradients can be noisy or saturated~\cite{simonyan2013deep}. To address this, several improvements have been proposed:

\begin{itemize}
    \item \textbf{Integrated Gradients}~\cite{sundararajan2017axiomatic}: integrates gradients from a baseline to overcome saturation.
    \item \textbf{Grad-CAM}~\cite{selvaraju2017grad}: highlights important regions using gradient-weighted feature maps.
    \item \textbf{SmoothGrad}~\cite{smilkov2017smoothgrad}: averages gradients across noisy perturbations to reduce variance.
\end{itemize}

Beyond post-hoc methods, some approaches incorporate saliency directly into training. Saliency-guided training~\cite{ismail2021improving} masks low-importance regions using
\[
x_{\text{masked}} = x \odot \mathcal{M}(S(x), \lambda),
\]
encouraging the model to rely on meaningful features. SMOOT~\cite{karkehabadi2024smoot} further adapts the masking fraction dynamically to balance interpretability and performance.

\subsection{Application to Genomic Sequences}

Compared to image models, saliency methods in genomics remain less explored. Genomic sequences pose unique challenges because they are discrete (A, T, G, C), structured, and biologically constrained. Genomic Benchmarks~\cite{gresova2022genomic} provides standardized datasets, but most work focuses on accuracy rather than interpreting \emph{why} models make specific predictions.
We adapt gradient-based analysis specifically for genomic sequences and introduce:

\begin{enumerate}
    \item nucleotide-type aggregation to reveal base-specific importance,
    \item positional importance statistics identifying recurrent high-impact regions,
    \item class-specific comparison distinguishing coding versus intergenic features.
\end{enumerate}

Together, these analyses provide interpretable insight into sequence classification and highlight biologically plausible patterns learned by the model.

\section{Methodology}

\subsection{Dataset and Task Formulation}

We study binary classification of coding versus intergenomic regions using the \texttt{demo\_coding\_vs\_intergenomic\_seqs} dataset from Genomic Benchmarks~\cite{gresova2022genomic}. The dataset contains 100{,}000 DNA sequences of fixed length \(L = 200\), evenly split between coding regions from human transcripts and non-coding intergenomic regions. We use a 75--25 train--test split. A sequence is represented as
\[
\mathbf{x} = [x_1, x_2, \ldots, x_L], \qquad x_i \in \{A, T, G, C, N\},
\]
and we learn a classifier \(f_\theta\) that predicts \(y \in \{0,1\}\) using the binary cross-entropy loss:
\[
\mathcal{L}(\theta)
=
-\mathbb{E}_{(x,y)}
\left[
y \log f_\theta(x) + (1-y)\log(1-f_\theta(x))
\right].
\]

\subsection{Model Architecture}




We use a compact 1D CNN consisting of an embedding layer that maps nucleotides to 128-dimensional vectors, followed by three convolutional layers (kernel size 8) with 32, 16, and 4 channels, each paired with batch normalization, ReLU activation, and max-pooling (stride 2) to extract hierarchical 8-mer features. The resulting \(\sim\!200\) flattened features are passed through a 512-unit fully connected layer with 0.3 dropout and finally a 2-logit output layer for binary classification.





\subsection{Gradient-Based Importance Analysis}

To interpret model predictions, we compute gradients with respect to the input embeddings.  
For a trained model \(f_\theta\) and target label \(y\):
\[
\mathbf{G}
=
\frac{\partial f_\theta(\mathbf{x})_y}{\partial \mathbf{E}(\mathbf{x})},
\]
where \(\mathbf{E}(\mathbf{x}) \in \mathbb{R}^{L \times d}\) is the embedding matrix (\(d=128\)).

We define position-level importance as:
\[
I_j = \sum_{k=1}^{d} |G_{j,k}|, \qquad j=1,\ldots,L.
\]

\begin{figure*}[ht!]
    \centering
    \includegraphics[width=0.72\linewidth]{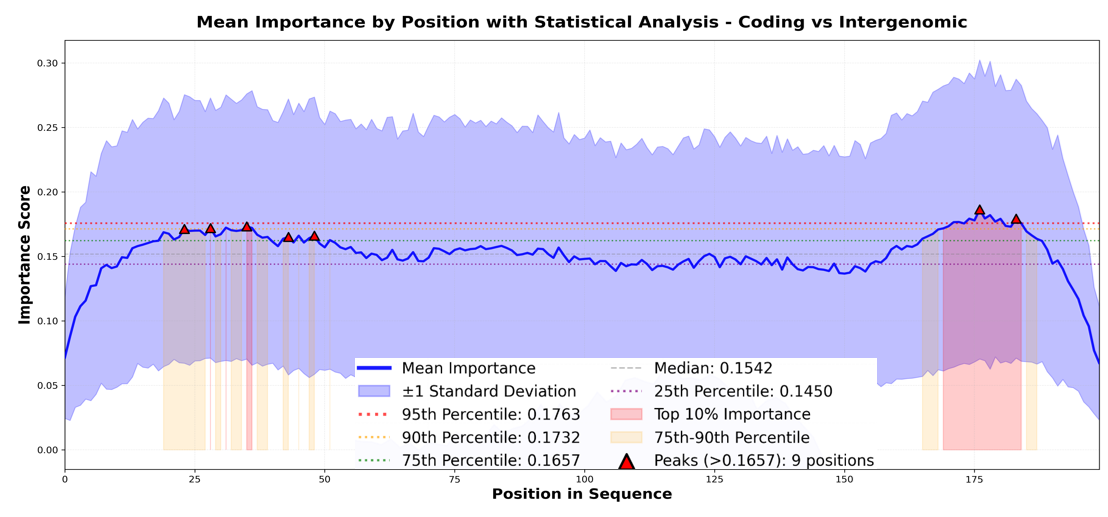}
    \caption{Positional mean importance with confidence bands. Nine positions exceed the 75th percentile threshold.}
    \label{fig:positional}
\end{figure*}

\begin{figure*}[h!]
    \centering
    \includegraphics[width=0.69\linewidth]{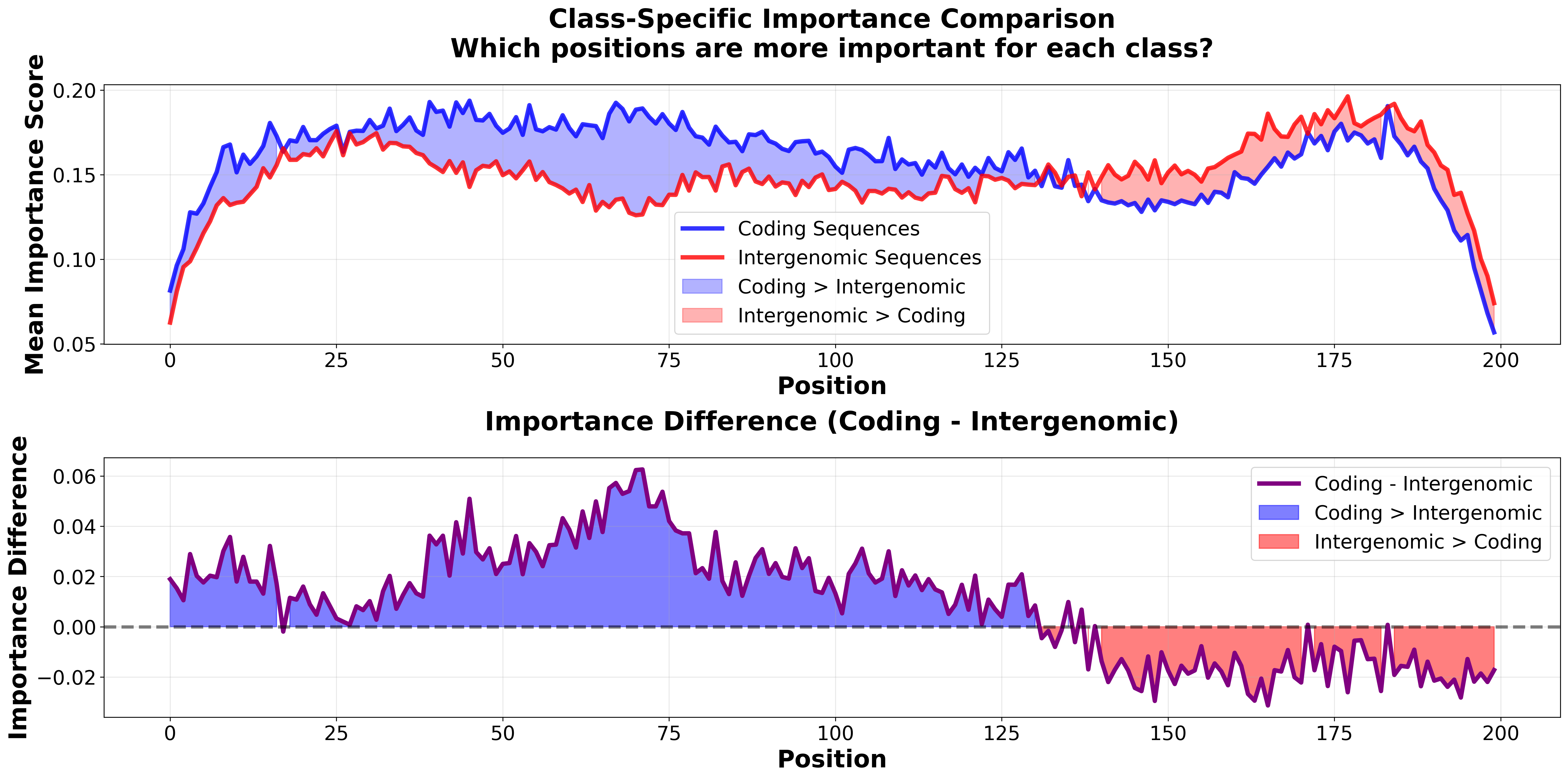}
    \caption{Coding vs.\ intergenomic importance patterns. Bottom: positional differences.}
    \label{fig:class}
\end{figure*}

\subsection{Positional Importance Analysis}

Across 1{,}000 sequences, we compute:

\[
\mu_j = \frac{1}{N} \sum_{i=1}^{N} I_{i,j},
\qquad
\sigma_j = 
\sqrt{
\frac{1}{N-1}
\sum_{i=1}^{N}(I_{i,j} - \mu_j)^2
}.
\]

Positions with
\[
\mu_j > \text{percentile}_{75}
\]
are considered high-importance hot spots; nine such positions are identified.

\subsection{Class-Specific Importance Comparison}

For each class \(c \in \{0,1\}\), we compute average importance vectors \(\boldsymbol{\mu}_c\) using 250 sequences.  
The difference
\[
\boldsymbol{\delta} = \boldsymbol{\mu}_0 - \boldsymbol{\mu}_1
\]
highlights coding-specific regions (\(\delta > 0\)) and intergenomic-specific regions (\(\delta < 0\)).

\section{Results}

\subsection{Model Performance}

Our \texttt{SimpleSaliencyCNN} achieves strong performance on the coding vs.\ intergenomic task.  
With only \textbf{91,145 parameters}, the model remains lightweight while providing accurate and interpretable predictions. We obtain a validation accuracy of \textbf{91.87\%}, outperforming baseline CNN models (Table~\ref{tab:acc}).

\begin{table}[h!]
\centering
\begin{tabular}{l c}
\hline
\textbf{Method} & \textbf{Validation Accuracy} \\
\hline
SimpleSaliencyCNN (Ours) & \textbf{91.87\%} \\
HyenaDNA & 91.3\% \\

\hline
\end{tabular}
\caption{Validation accuracy comparison.}
\label{tab:acc}
\end{table}

\begin{figure*}[h!]
    \centering
    \includegraphics[width=1 \linewidth]{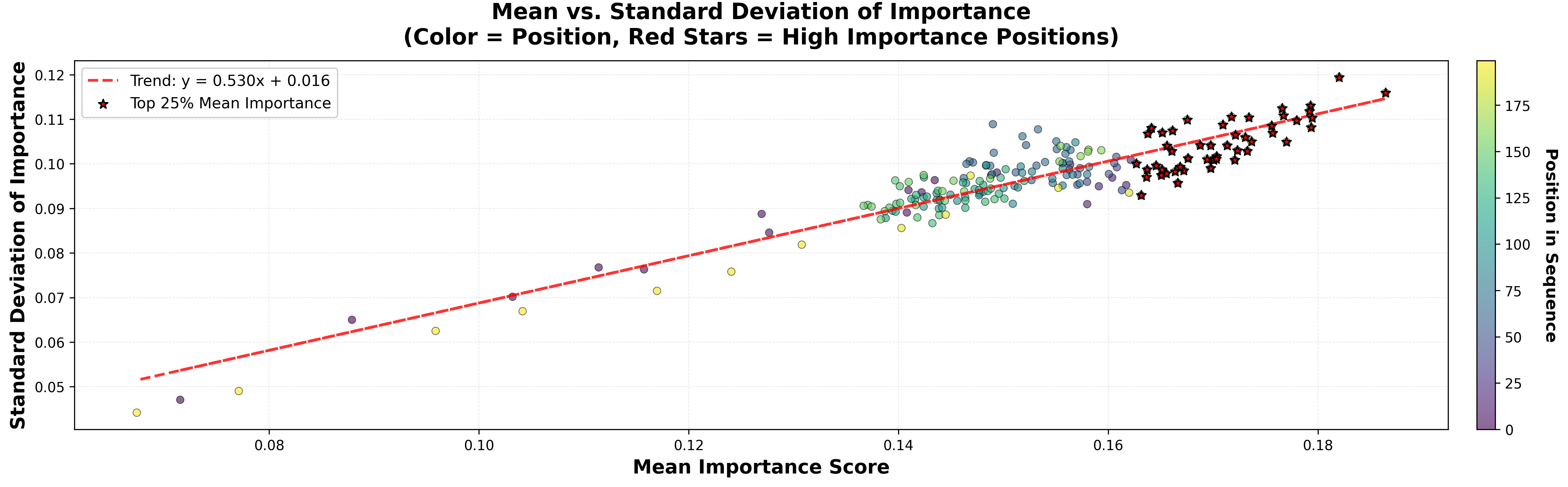}
    \caption{Mean vs.\ standard deviation of positional importance. Red stars highlight top 25\% positions.}
    \label{fig:cov}
\end{figure*}
\vspace{8 pt}

\subsection{Extended Gradient-Based Analysis}

We compute gradient-based importance scores across 1,000 sequences to identify which positions most influence predictions. The positional importance profile (Figure~\ref{fig:positional}) reveals two consistent high-impact regions at \textbf{20--40} and \textbf{160--190}. Class-specific comparisons (Figure~\ref{fig:class}) show that coding sequences concentrate importance in \textbf{20--100}, whereas intergenomic sequences emphasize \textbf{150--190}. The relationship between mean and variance (Figure~\ref{fig:cov}) exhibits a positive correlation (\(r = 0.530\)), indicating that the most influential positions also show stronger context-dependent variation.

\subsection{Biological Interpretation}

These patterns align with known genomic structure: periodic importance suggests implicit learning of reading-frame signals; high-impact peaks match typical promoter/enhancer locations; and the distinct importance ranges for coding and intergenomic sequences reflect expected GC-rich codon patterns versus regulatory motif signatures.\\

\section{Conclusion}
This work demonstrates that a lightweight CNN can achieve strong genomic sequence classification performance while providing interpretable, position-level evidence for its predictions. With 91.87\% validation accuracy, our model identifies nine prominent importance peaks and reveals distinct attribution patterns between coding (positions 20--100) and intergenomic (positions 150--190) sequences. The observed mean--variance correlation (\(r = 0.530\)) further suggests that highly influential positions exhibit greater variation across samples, highlighting the context dependence of learned sequence features. By combining positional statistics with class-specific gradient analysis, our framework moves beyond predictive accuracy to characterize where a sequence model concentrates its attention. These results provide a practical basis for generating testable biological hypotheses, while motivating future validation against annotated genomic elements and independent datasets.\\

{
    \small
    \bibliographystyle{ieeenat_fullname}
    \bibliography{main}
}

\end{document}